\documentclass{article}
\usepackage{spconf,amsmath,amssymb,graphicx}
\usepackage{enumitem}
\setlist{nosep, leftmargin=14pt}
\usepackage{nicematrix,tikz,booktabs}
\usepackage{dsfont}
\newlength{\tempdima}
\usepackage{multirow}
\newcommand{\rowname}[1]{\rotatebox{90}{\makebox[\tempdima][c]{#1}}}
\usepackage{enumitem}
\usepackage[export]{adjustbox}
\usepackage{graphbox}

\usepackage[numbers, sort&compress]{natbib}

\title{DSA-CycleGAN: A Domain Shift Aware CycleGAN for Robust Multi-Stain Glomeruli Segmentation}
\name{Zeeshan Nisar$^{\star}$ \quad Friedrich Feuerhake$^{\dagger}$ \quad Thomas Lampert$^{\star}$}
\address{$^{\star}$ICube, University of Strasbourg, France \\ 
$^{\dagger}$Institute of Pathology, Hannover Medical School, Germany}
\begin{document}
%
\maketitle
\begin{abstract}
A key challenge in segmentation in digital histopathology is inter- and intra-stain variations as it reduces model performance. Labelling each stain is expensive and time-consuming so methods using stain transfer via CycleGAN, have been developed for training multi-stain segmentation models using labels from a single stain. Nevertheless, CycleGAN tends to introduce noise during translation because of the one-to-many nature of some stain pairs, which conflicts with its cycle consistency loss. To address this, we propose the Domain Shift Aware CycleGAN, which reduces the presence of such noise. Furthermore, we evaluate several advances from the field of machine learning aimed at resolving similar problems and compare their effectiveness against DSA-CycleGAN in the context of multi-stain glomeruli segmentation. Experiments demonstrate that DSA-CycleGAN not only improves segmentation performance in glomeruli segmentation but also outperforms other methods in reducing noise. This is particularly evident when translating between biologically  distinct stains. The code is publicly available at https://github.com/zeeshannisar/DSA-CycleGAN.
\end{abstract}


\section{Introduction} 
\label{sec:intro}
Histopathology, the microscopic examination of tissues and cells, remains the gold standard for diagnosing a wide range of diseases. The digitisation of histopathological slides to whole slide images (WSIs) has led to a new research field known as `digital histopathology'. This has significantly expanded histopathology's scope  by enabling remote diagnostics, enhancing collaboration among pathologists, and supporting the development of state-of-art deep learning (DL) based computer aided diagnostic (CAD) solutions \cite{williams2024digital}. 

WSI datasets often consist of consecutive slides stained differently, revealing different cell types and tissues structures. As each stain highlights different features, even consecutive WSIs that represent the same anatomical structures appear differently, leading to inter-stain variations. Furthermore, the staining procedure itself is prone to variability due to inter-subject variations, laboratory specific techniques, and scanner characteristics, leading to intra-stain variations \cite{vasiljevic2022cyclegan}. 

To enable accurate diagnosis and disease characterisation, pathologists often analyse and integrate information from various stains, e.g.\ it is important to focus on specific structures of interest, such as glomeruli, in multiple stains in the context of kidney analysis as they play a critical role in diagnosing pathologies like kidney allograft rejection. To automate such an analysis, glomeruli must be detected (segmented) in each differently stained tissue section. However, inter- and intra-stain variations cause significant drops in DL models' performance when applied to out-of-sample stains, even for the same task \cite{vasiljevic2021towards}. 
A straightforward but costly, solution is to acquire sufficient labels for each stain and to train separate DL models. However, this approach is impractical and inefficient.

To overcome this, researchers have focused on unsupervised domain adaptation based stain transfer \cite{vasiljevic2021towards} using unpaired image-to-image translation frameworks, such as CycleGAN \cite{zhu2017unpaired}. This enables the development of single model multi-stain segmentation algorithms with labels from only one (source, S) stain, eliminating the need of labels in other (target, T) stains. Notably, Multi-Domain Supervised 1 (MDS1) \cite{gadermayr2019generative} allows a segmentation model to be applied to a target stain by translation to the source stain (T$\rightarrow$S). \citep{lo2020cycleconsistent} followed suite by introducing a new approach to translation and inspired by their successes, stain augmentation based solutions \cite{vasiljevic2021towards, vasiljevic2023histostargan} have also been proposed to train a single stain-invariant segmentation model using labels from only the source stain. These multi-stain segmentation methods have shown great potential when applied to unlabelled target stains, achieving almost fully-supervised performance \cite{gadermayr2019generative, vasiljevic2021towards}. 

Nevertheless, these methods are limited by the noise produced during stain transfer, leading to domain shift (DS) in the translated stains (images) \cite{nisar2022towards}. This DS is particularly pronounced when translating between information-rich and information-poor domains \cite{vasiljevic2021towards}. Information-poor domains being those that highlight or mark a limited set of biological features, such as immunohistochemical (IHC) stains, whereas information-rich domains, i.e.\ histochemical (HC) stains, provide a broader representation of biological features \citep{abdo2024stairwaytostain}. IHC markers are generally difficult to infer from HC stains \cite{vasiljevic2021self}, so additional information is needed for  the CycleGAN model to reconstruct them during cycle-consistency \cite{wang2018cyclegan, bashkirova19, vasiljevic2022cyclegan, nisar2022towards}, which takes the form of 
imperceptible noise  in the translated images. \citep{nisar2022towards} showed how to detect and measure this noise in the real source and T$\rightarrow$S translated stains using the Domain Shift Metric (DSM) \cite{stacke2020measuring}, which has strong negative correlation with MDS1 multi-stain segmentation performance.

Building on this, DSM is used to mitigate the impact of such noise during CycleGAN translation. 
As such, we propose the domain shift aware CycleGAN (DSA-CycleGAN), a self-guided strategy for learning translations with minimal domain shift (noise), which improves stain transfer, thereby enhancing the performance of 
multi-stain segmentation.

In general computer vision literature several other methods have been introduced to reduce such noise. 
Particularly, \citep{wang2018cyclegan} and \citep{shao2021spatchgan} proposed to attenuate the cycle-consistency constraint to avoid such noise 
 and \citep{nizan2019breaking} suggested to remove it completely. \citep{bashkirova19} proposed to add random Gaussian noise to the translated image, prior to reconstructing the original. This disrupts the embedded information, increasing the reconstruction error and forcing the generator to learn more accurate translations. \citep{chu2017cyclegan} proposed to separate the noise into an additional image channel during CycleGAN training. Of the above, \citep{shao2021spatchgan} and \citep{nizan2019breaking} are designed for applications involving significant geometric changes and shape deformations between the source and target domains, which would invalidate any ground truth data associated with the source image, it is therefore not appropriate in this application. Finally, in a model specific to histopathology, \citep{bouteldja2022improving} suggested to integrate a pre-trained source stain segmentation model into CycleGAN in a self-supervised way to improve stain transfer. A similar approach is proposed by \citep{Bao23} that relies on cell segmentation using a model pre-trained on multiplexed tissue samples.

While such approaches have improved upon the original CycleGAN \cite{zhu2017unpaired}, most are designed for computer vision applications and have never been applied to histopathology related tasks. Therefore, in this study we propose to use the methods presented by  \citep{bashkirova19}, \citep{chu2017cyclegan}, and \citep{bouteldja2022improving} for two purposes: (1) to provide a comparative analysis to DSA-CycleGAN; and (2) to investigate their effectiveness for enhancing multi-stain glomeruli segmentation, which to the best of our knowledge, has not previously been explored. \citep{bouteldja2022improving} is the only method specifically developed for histopathology and is conceptually most similar to DSA-CycleGAN. Nevertheless, it relies upon comparing the consistency of segmentation masks during translation using a pre-trained source stain segmentation model. This limits its loss' application to images that originate from the source domain, i.e.\ the product of the CycleGAN's cycle consistency and identity transforms. 
DSA-CycleGAN avoids this as it does not rely on segmentation masks during CycleGAN training, but instead measures data distribution differences (relative to the source images). This allows it to be applied to the previously mentioned CycleGAN components, in addition to the T$\rightarrow$S translations.

\section{Methods}
\label{methods}
\begin{figure}[tb]
  \centering
  \includegraphics[width=\linewidth]{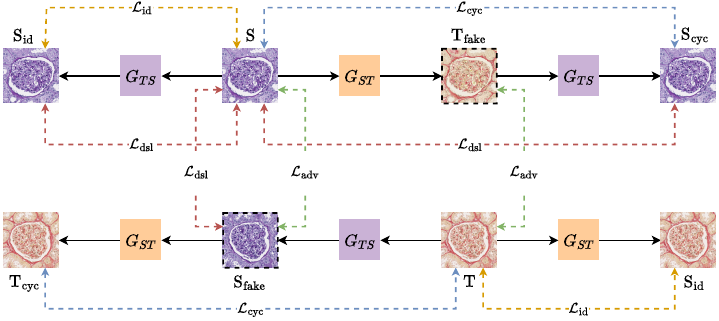}
  \caption{DSA-CycleGAN: an additional loss term,  $\mathcal{L}_{\text{dsl}}$, is incorporated into the CycleGAN architecture, while the original components (adversarial, $L_\textrm{adv}$, cycle-consistency, $L_\textrm{cyc}$, and identity mapping, $L_\textrm{id}$, losses) remain unchanged. $\mathcal{L}_{\text{dsl}}$ guides features to be  more closely aligned with the source domain.}
  \label{figure:dsa_cyclegan}
\end{figure}

Before presenting the DSA-CycleGAN, an overview of the domain shift metric is given. The DSM \cite{stacke2020measuring} measures the difference between two domains' distributions using a pre-trained model's feature representations. Consider a CNN with layers $\{l_1,\dots,l_L\}$. Let $\Phi(x) = \{\phi_{l1}(x), \dots, \phi_{lk}(x)\}$, such that $\Phi_{lk}(x) \in \{\mathbb{R}^{h \times w}\}$, denote the filter activations at layer $l$ and filter $k$. The mean value of each $\Phi_{lk}(x)$ is calculated as
\begin{equation} 
    c_{lk}(x) = \frac{1}{h w} \sum_{i,j}^{h,w} \Phi_{lk}(x)_{i,j}.
	\label{eq:domain_shift_part1}
\end{equation}

Let $p^S_{c_{lk}}$ denote a distribution of $c_{lk}(x)$, for each source domain ($S$) image, $x$, and $p^{S'}_{c_{lk}}$ denote the same for the translated T$\rightarrow$S domain ($S'$), DSM is then defined as
\begin{equation} 
	\text{DSM}(p^S, p^{S'}) = \frac{1}{k} \sum_{i=1}^{k} \mathcal{W} \left( p^S_{c_{lk}}, p^{S'}_{c_{lk}} \right),
	\label{eq:domain_shift_part2}
\end{equation}
where $\mathcal{W}$ is the Wasserstein distance \cite{ramdas2015onwt} between $p^S_{c_{lk}}(x)$ and $p^{S'}_{c_{lk}}(x)$, which tends towards 0 when $S$ and $S'$ are similar. 

The original CycleGAN architecture can then be modified by integrating the DSM, Eq.\ \eqref{eq:domain_shift_part2}, as a loss ($\mathcal{L}_{\text{dsl}}$) to minimise the impact of domain shift in the translated images. Because DSM measures this domain shift using a pretrained source segmentation model, without requiring ground truth during CycleGAN training, it can be incorporated into any part of the CycleGAN that results in source-like images. 
Specifically, the T$\rightarrow$S translations ($\mathcal{L}_{\text{dsl\_fake}}$) in addition to the source cycle consistency reconstructions ($\mathcal{L}_{\text{dsl\_cyc}}$) and source identity mappings ($\mathcal{L}_{\text{dsl\_id}}$), as visualised in Fig.\ \ref{figure:dsa_cyclegan}, such that 
\begin{align*}
	\mathcal{L}_{\text{dsl}} &=  \mathcal{L}_{\text{dsl\_fake}} + \mathcal{L}_{\text{dsl\_cyc}} + \mathcal{L}_{\text{dsl\_id}} \nonumber \\
	&= \mathds{E}_{s \sim S}\mathds{E}_{t \sim T} [\text{DSM}(p^{s}, p^{G_{TS}(t)}) \nonumber \\
	&\quad + \text{DSM}(p^{s}, p^{G_{TS}(G_{ST}(s))}) + \text{DSM}(p^{s}, p^{G_{TS}(s)})].
\end{align*}
This results in the following CycleGAN loss function
\begin{align*}
	\mathcal{L}&_{\text{CycleGAN}}(G_{ST},G_{TS},D_S,D_T) =  \mathcal{L}_{\text{adv}}(G_{ST},D_T,G_{TS},D_S) \nonumber \\
	&\quad + w_{\textrm{cyc}} \mathcal{L}_{\text{cyc}}(G_{ST},G_{TS}) + w_{\textrm{id}} \mathcal{L}_{\text{id}}(G_{ST},G_{TS}) \nonumber \\
	&\quad + w_{\textrm{dsl}}\mathcal{L}_{\text{dsl}}(G_{ST},G_{TS}).
\end{align*}

\section{Experimental Setup}
\label{experiment_results}
The models were tested on  multi-stain glomeruli segmentation using  MDS1 \cite{gadermayr2019generative}. A segmentation model, such as UNet \cite{ronneberger2015u}, is trained on a labelled source stain (PAS in our case). Next, target stain test images (herein Jones H\&E, Sirius Red, CD68, CD34) are translated (T$\rightarrow$PAS) using a stain translation model. Finally the source-trained segmentation model ($\text{UNet}_{\text{PAS}}$) is applied to the T$\rightarrow$PAS images to obtain their segmentation masks. The UNet and translation model training is detailed below. 
To account for random variations, each CycleGAN was trained 3 times per target stain, and the UNet was trained 5 times for each CycleGAN, resulting in 15 experimental repetitions per target stain. To evaluate the impact of $\mathcal{L}_{\text{dsl}}$, we also conduct an ablation.
 Specifically: (1) the effect of applying only $\mathcal{L}_{\text{dsl\_fake}}$, referred to as ``Ours ({dsl\_fake})''; (2) the combined application of $\mathcal{L}_{\text{dsl\_cyc}}$ and $\mathcal{L}_{\text{dsl\_id}}$, ``{Ours} ({dsl\_cyc, dsl\_id})''; and (3) the impact of integrating all components, DSA-CycleGAN. 

The following models are also evaluated: (a) CycleGAN with Extra-channels (w/Extra-channels) \citep{chu2017cyclegan}; (b) CycleGAN with Gaussian-noise (w/Gaussian-noise) \citep{bashkirova19}; and (c) CycleGAN with Self-supervision (w/Self-supervision) \citep{bouteldja2022improving}. 
\begin{description}[leftmargin=0pt,topsep=0pt,noitemsep]
\item[Dataset: ]The dataset was provided by the Hannover Medical School, containing renal tissue samples collected from $10$ patients who had tumor nephrectomy for renal carcinoma (tissue was selected as far as possible from the tumors to represent normal renal glomeruli). Using an automated staining tool, Ventana Benchmark Ultra, 3µm thick sections were taken from paraffin-embedded samples and stained with PAS, Jones H\&E, Sirius Red, and two IHC markers (CD68, and CD34). An Aperio AT2 scanner was used at $40$ magnification, $0.253$ m/pixel. Pathology specialists annotated the glomeruli through Cytomine \citep{maree2016collaborative}. The dataset was split into $4$ train, $2$ valid., and $4$ test patients, with glomeruli counts of: PAS - 662 (train), 588 (valid), 1092 (test); Jones H\&E - 590 (valid), 1043 (test); Sirius Red - 576 (valid), 1049 (test); CD34 - 595 (valid), 1019 (test); CD68 - 521 (valid), 1046 (test). 

\item[Evaluation Metrics: ] 
Evaluating stain transfer is challenging due to the lack of ground-truth, 
\cite{bouteldja2022improving} and \cite{dohmen2025similarity} instead propose to use a downstream segmentation task, i.e.\ MDS1. As such, we use the F$_1$ score to compare the segmentation performance of a pretrained UNet model (trained on PAS) across various CycleGAN based T$\rightarrow$PAS translated stains. 

\item[Training Details: ]
$512\times512$ patches were used as they encompass glomeruli and part of the surrounding tissue at the level-of-detail used. The UNet was trained as in \cite{vasiljevic2021towards} and   
the original CycleGAN implementation \cite{zhu2017unpaired} were used. 
For CycleGAN w/Gaussian-noise, $\sigma=0.0125$ was found (from 0.0125, 0.025, 0.05, 0.1, 0.5, 0.9) to be best across all target stains, $w_{\text{dsl}}$ (DSA-CycleGAN) and $w_{\text{seg}}$ (CycleGAN w/Self-supervision) were found to be 1.0 (from 1.0, 5.0, 10.0). 
\end{description}

\section{Results} 
\label{results}

\begin{table*}[tb]
	\centering
	\footnotesize
	\caption{F$_1$ scores of $\text{UNet}_{\text{PAS}}$ applied to full test slides of various T$\rightarrow$PAS translated stains using different CycleGAN 
	models. 
	The highest F$_1$ scores are indicated in bold and are statistically significant from the baseline CycleGAN ($p<0.05$).}
	\label{table-results}
	\begin{NiceTabular}{ccc:c:cc:c}
		\toprule
		\Block{3-1}{Approach} & \Block{1-6}{Test Stains} \\
		\cmidrule{2-7}
		& \Block{1-3}{HC Stains} & & & \Block{1-3}{IHC Stains} & & \\
		\cmidrule{2-7}
		& Jones H\&E & Sirius Red & Average & CD68 & CD34 & Average \\
		\midrule
		\midrule
		CycleGAN (Baseline) & 0.844 \scriptsize{(0.026)} & 0.860 \scriptsize{(0.023)} & 0.852 \scriptsize{(0.025)} & 0.643 \scriptsize{(0.031)} & 0.747 \scriptsize{(0.021)} & 0.695 \scriptsize{(0.026)} \\
		\midrule
				
		w/Extra-channels & 0.862 \scriptsize{(0.019)} & 0.871 \scriptsize{(0.020)} & 0.867 \scriptsize{(0.020)} & 0.634 \scriptsize{(0.037)} & 0.669 \scriptsize{(0.041)} &
		0.652 \scriptsize{(0.039)} \\
		
		w/Gaussian-noise & \textbf{0.865} \scriptsize{(0.016)} & \textbf{0.878} \scriptsize{(0.015)} & \textbf{0.872} \scriptsize{(0.016)} & 0.669 \scriptsize{(0.026)} & 0.749 \scriptsize{(0.028)} & 0.709 \scriptsize{(0.027)} \\
		
		w/Self-supervision & 0.840 \scriptsize{(0.027)} & 0.866 \scriptsize{(0.021)} & 0.853 \scriptsize{(0.024)} & 0.686 \scriptsize{(0.020)} & 0.753 \scriptsize{(0.024)} & 0.720 \scriptsize{(0.022)} \\
		\midrule
		
		Ours (dsl\_fake) & 0.851 \scriptsize{(0.020)} & 0.859 \scriptsize{(0.024)} & 0.855 \scriptsize{(0.022)} & 0.652 \scriptsize{(0.028)} & 0.735 \scriptsize{(0.024)} & 0.694 \scriptsize{(0.026)} \\
		
		Ours (dsl\_cyc, dsl\_id) & 0.860 \scriptsize{(0.017)} & 0.856 \scriptsize{(0.023)} & 0.858 \scriptsize{(0.020)} & 0.677 \scriptsize{(0.028)} & 0.724 \scriptsize{(0.028)} & 0.701 \scriptsize{(0.028)} \\
		
		DSA-CycleGAN & 0.855 \scriptsize{(0.019)} & 0.867 \scriptsize{(0.021)} & 0.856 \scriptsize{(0.023)} &\textbf{0.694} \scriptsize{(0.021)} & \textbf{0.763} \scriptsize{(0.012)} & \textbf{0.730} \scriptsize{(0.017)}\\
		\bottomrule
	\end{NiceTabular}
\end{table*}

\begin{figure}[tb] 
	\centering \scriptsize
	\setlength{\tabcolsep}{4pt}
	\settoheight{\tempdima}{\includegraphics[width=0.15\linewidth]{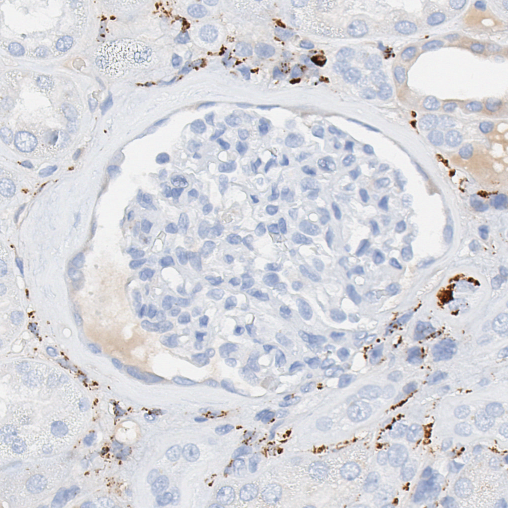}}
	\begin{tabular}{c@{ }c@{ }c@{ }c@{ }c@{ }c@{ }c@{ }}
        & \multirow{ 2}{*}{Target Stain} & \multirow{2}{*}{Baseline} & w/Extra- & w/Gaussian- & w/Self- & DSA- \\
        & & & channels & noise & supervision & CycleGAN \\
		\rowname{CD68} &  
		\includegraphics[width=0.15\linewidth]{16_real.png} &
		\includegraphics[width=0.15\linewidth]{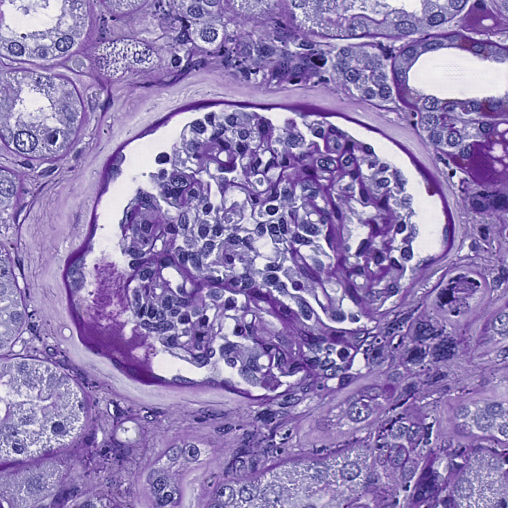} &
		\includegraphics[width=0.15\linewidth]{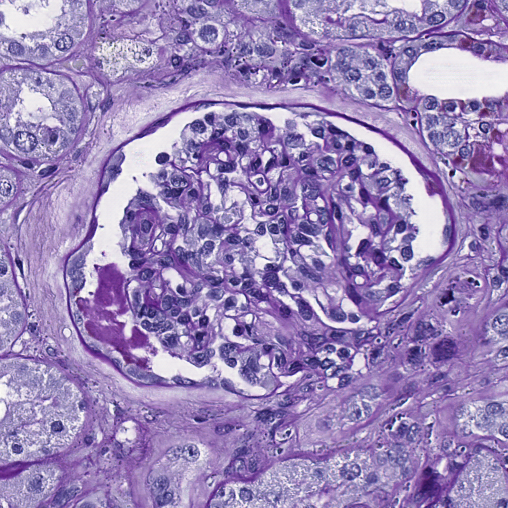} &
		\includegraphics[width=0.15\linewidth]{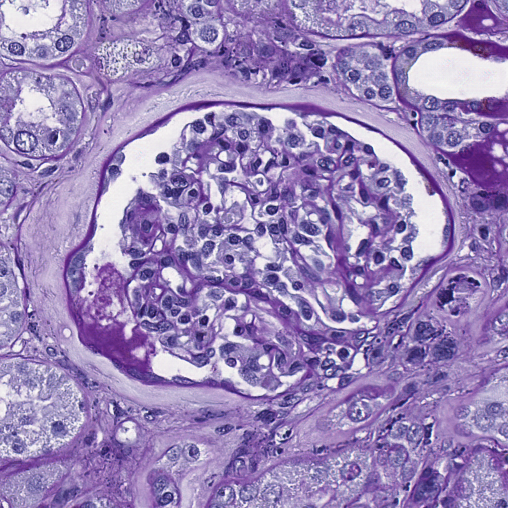} &
		\includegraphics[width=0.15\linewidth]{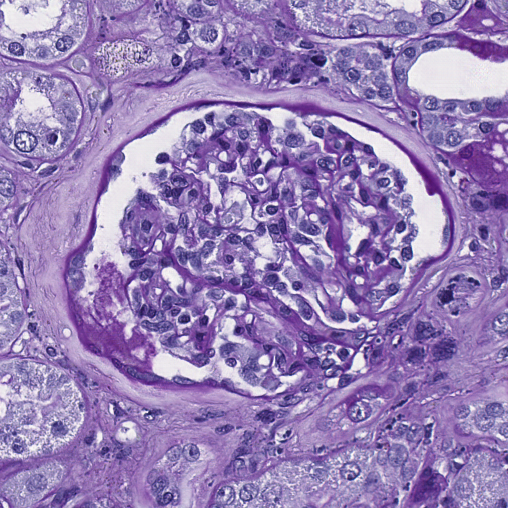} &
		\includegraphics[width=0.15\linewidth]{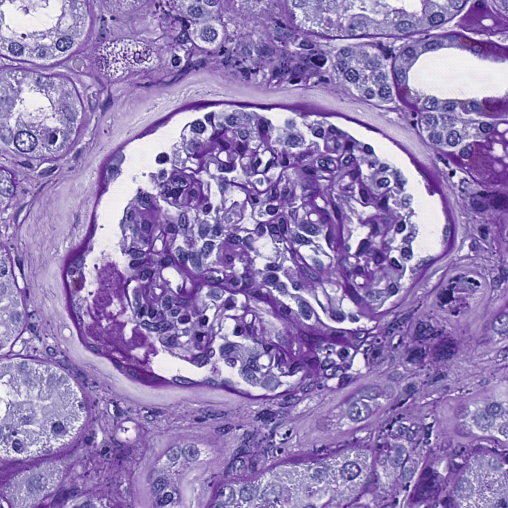}
		\\
		
		 & \multicolumn{1}{r}{\rowname{Predictions}} &
		\includegraphics[width=0.15\linewidth]{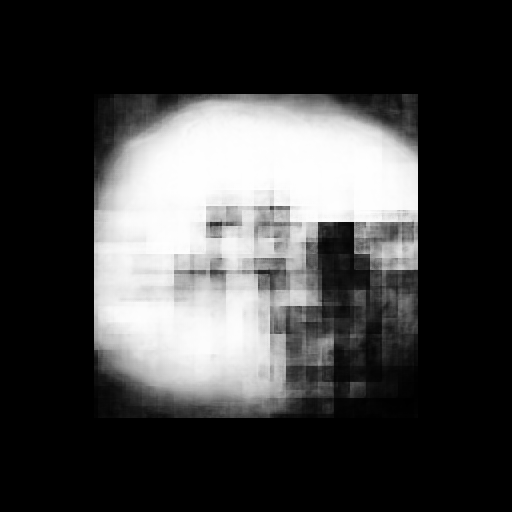} &
		\includegraphics[width=0.15\linewidth]{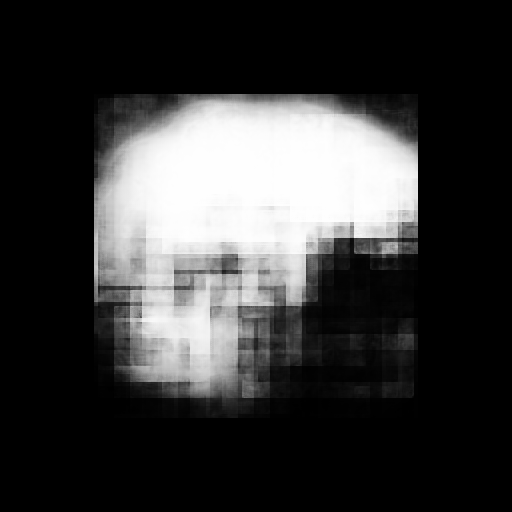} &
		\includegraphics[width=0.15\linewidth]{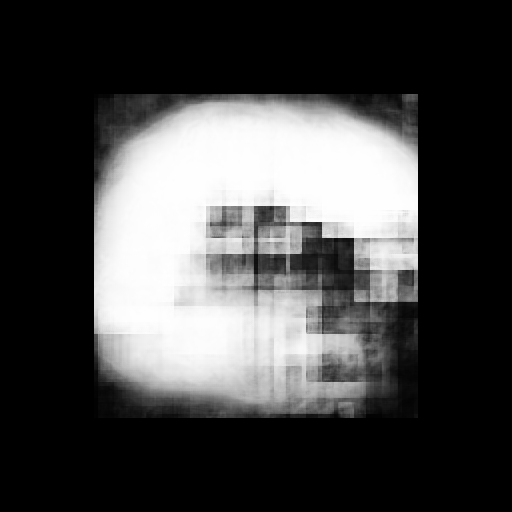} &
		\includegraphics[width=0.15\linewidth]{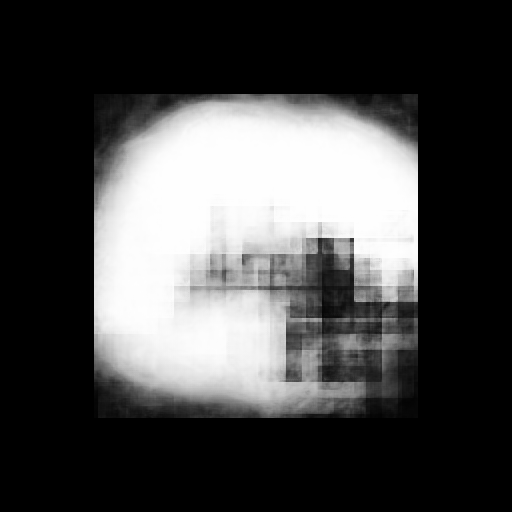} &
		\includegraphics[width=0.15\linewidth]{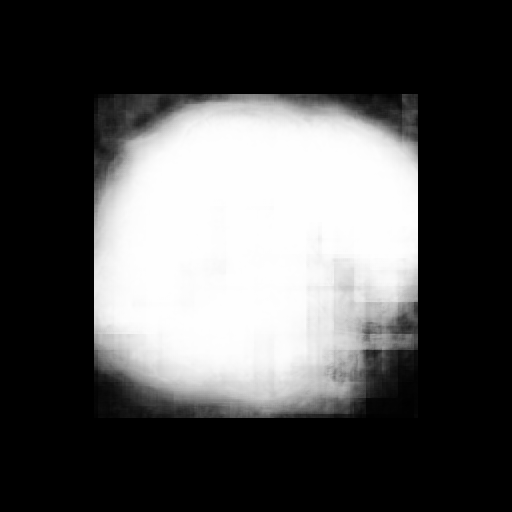}
		\\
		
		\rowname{CD34} &  
		\includegraphics[width=0.15\linewidth]{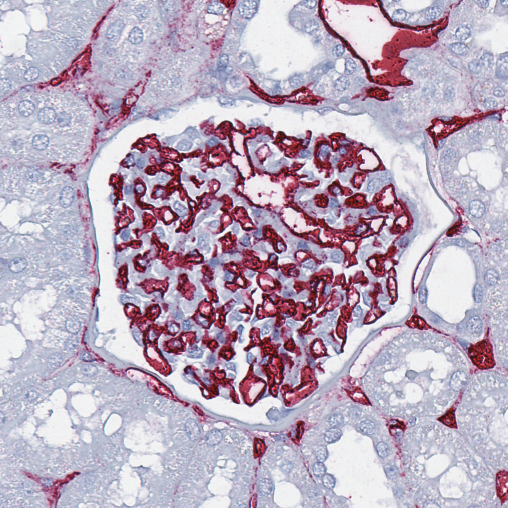} &
		\includegraphics[width=0.15\linewidth]{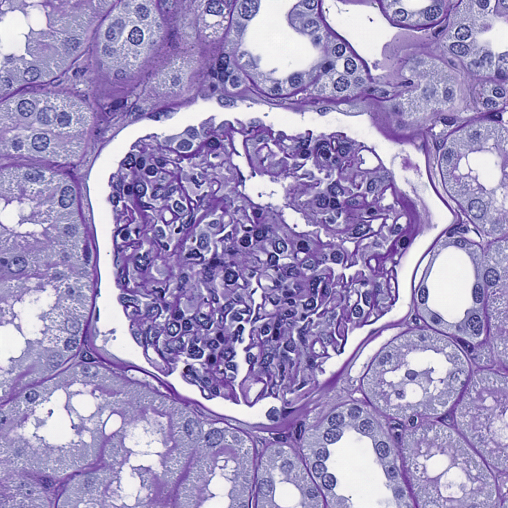} &
		\includegraphics[width=0.15\linewidth]{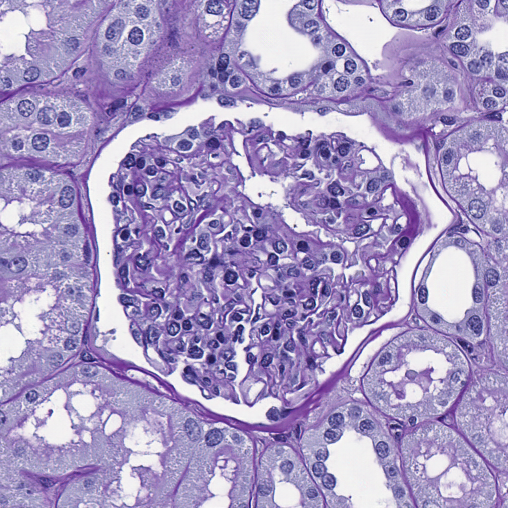} &
		\includegraphics[width=0.15\linewidth]{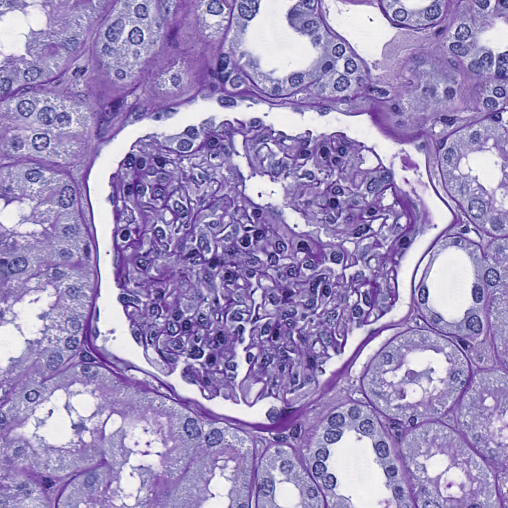} &
		\includegraphics[width=0.15\linewidth]{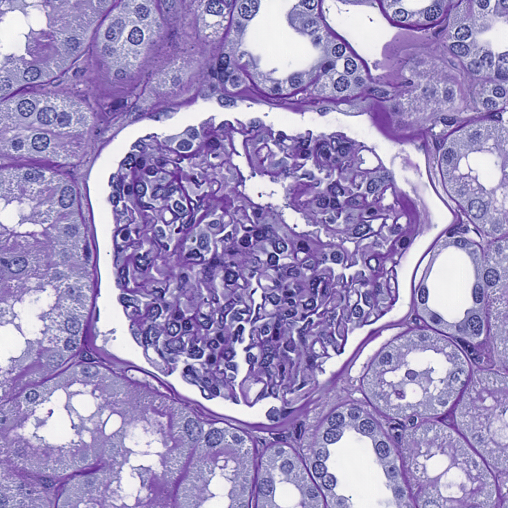} &
		\includegraphics[width=0.15\linewidth]{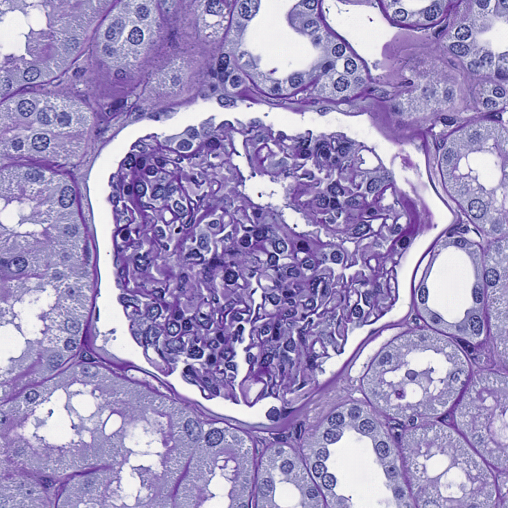}
		\\
		
		& \multicolumn{1}{r}{\rowname{Predictions}} &
		\includegraphics[width=0.15\linewidth]{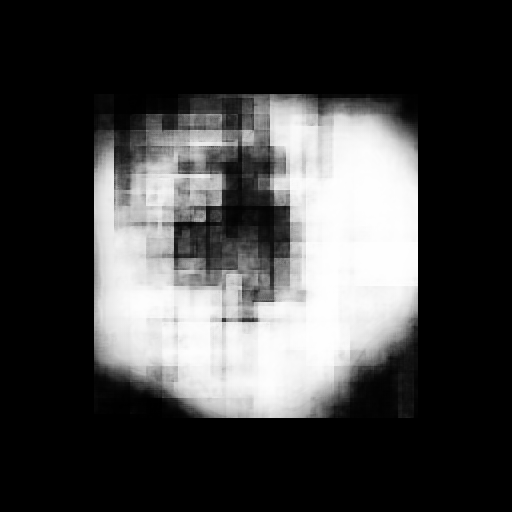} &
		\includegraphics[width=0.15\linewidth]{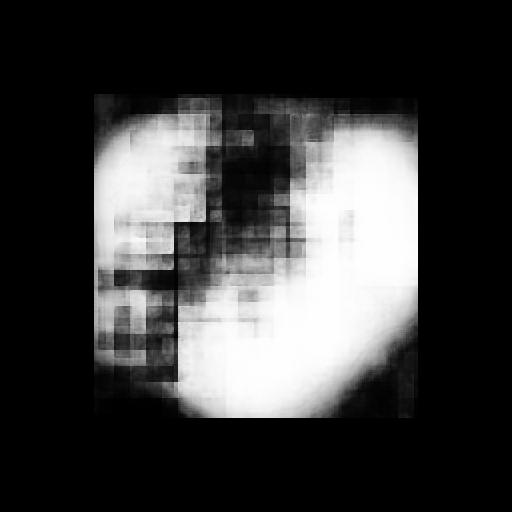} &
		\includegraphics[width=0.15\linewidth]{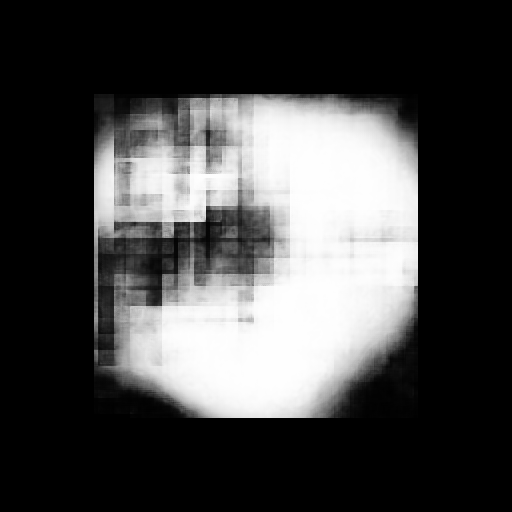} &
		\includegraphics[width=0.15\linewidth]{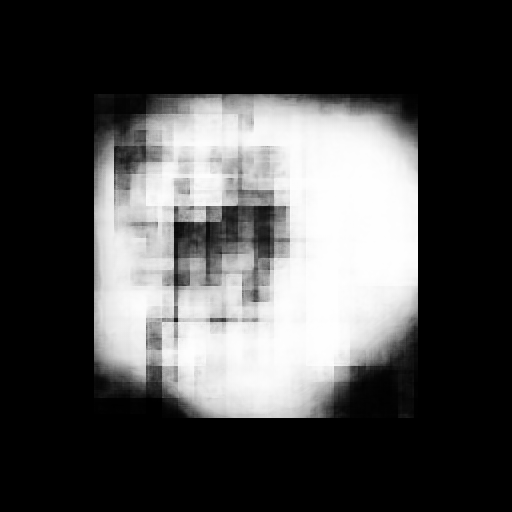} &
		\includegraphics[width=0.15\linewidth]{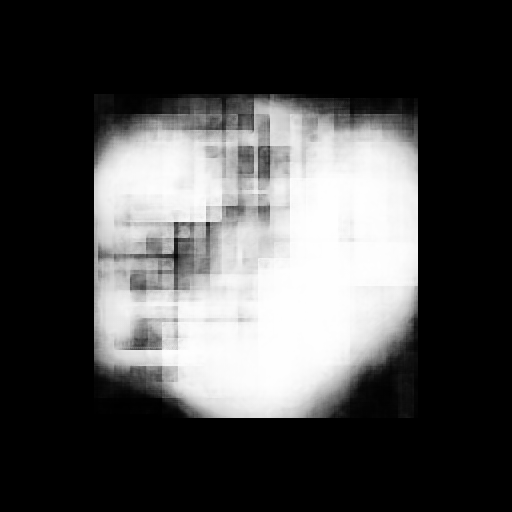}
	\end{tabular} 
	\caption{IHC target patches translated to PAS (T$\rightarrow$PAS). While all translations are visually realistic, differences are observed in the segmentation maps given by a pretrained PAS segmentation model ($\text{UNet}_{\text{PAS}}$), reflecting the numerical results, DSA-CycleGAN offers the most robust segmentation maps.} 
	\label{figure:stain-transfer}
\end{figure}

Table \ref{table-results} presents the results of multi-stain glomeruli segmentation.. For HC stains, all match or outperform the Baseline, which is more pronounced when using CycleGAN w/Gaussian-noise. For IHC stains, particularly CD68, more substantial improvements over the Baseline are observed except with CycleGAN w/Extra-channels. The original CycleGAN struggles with IHC stains as they are more biologically distinct from the source (PAS), requiring more noise to be introduced (giving lower PAS baseline performance). The proposed methods manage (to some extent) to mitigate this noise from the T$\rightarrow$PAS translations, where DSA-CycleGAN leads. 
The ablation study in the bottom of Table \ref{table-results} shows that all DSA-CycleGAN loss components are required, and that constraining the T$\rightarrow$S translation increases performance.

\begin{table*}[tb]
	\caption{F$_1$ scores of $\text{UNet}_{\text{PAS}}$ applied to 200 randomly selected CD68 stain samples from the KidneyArtPathology dataset. 
	The highest F$_1$ score is in bold and is statistically significant from the baseline CycleGAN ($p<0.05$).}
	\label{kidneyartpathology_scores}
	\centering
	\footnotesize
	\begin{NiceTabular}{ccccc}
		\toprule
		 CycleGAN (Baseline) & w/Extra-channels & w/Gaussian-noise & w/Self-supervision & DSA-CycleGAN\\
		\midrule
		\midrule
		0.877 \scriptsize{(0.025)} & 0.778 \scriptsize{(0.037)} & 0.886 \scriptsize{(0.020)} & 0.879 \scriptsize{(0.016)} & \textbf{0.891} \scriptsize{(0.013)} \\
		\bottomrule
	\end{NiceTabular}
\end{table*}

To show that noise resilience generalises, 
we randomly selected 200 glomeruli and negative CD68 patches from the public KidneyArtPathology dataset \citep{vasiljevic2023histostargan}. Table \ref{kidneyartpathology_scores} shows that all methods, except CycleGAN w/Extra-Channels,  outperform the Baseline with DSA-CycleGAN leading. Likely because it directly affects the target to source translation path.

\section{Discussion} 
\label{discussions}

Although the T$\rightarrow$PAS translations of different CycleGAN models appear realistic, Fig.\ \ref{figure:stain-transfer}, segmentation performance varies across stains and models. Notably, CycleGAN w/Gaussian-noise gave superior performance for HC stains, however, this is less pronounced for IHC stains, where CycleGAN w/self-supervision and DSA-CycleGAN lead. This can be attributed to CycleGAN w/Gaussian-noise's augmentation technique, which is particularly robust in handling low-amplitude perturbations \citep{bashkirova19}, such as in 
the biologically similar HC$\leftrightarrow$HC translation.  HC$\leftrightarrow$IHC translation is more biologically distant, requiring higher-amplitude perturbations, which are not effectively addressed by such augmentation.

\begin{figure}[tb] 
    \centering \scriptsize
    \setlength{\tabcolsep}{2pt}
    \begin{tabular}{cccc} 
        CD68$_{\text{real}}$ & CD68$\rightarrow$PAS & Extra-channel & CD68$_{\text{reconstructed}}$
        \\
        \includegraphics[width=0.22\linewidth]{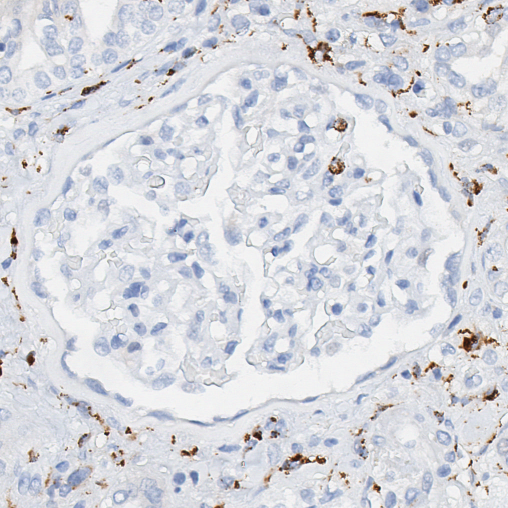} & 
        \includegraphics[width=0.22\linewidth]{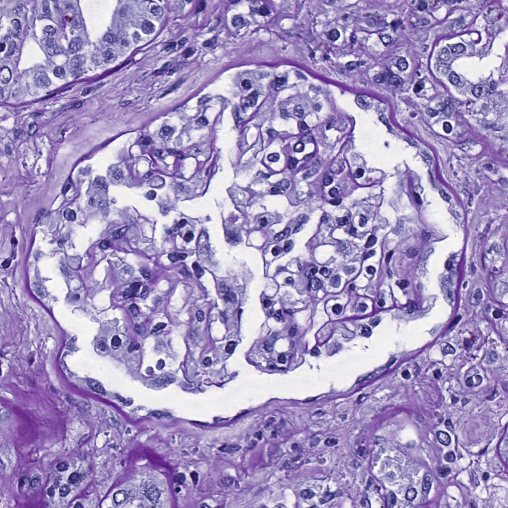} &
        \includegraphics[width=0.22\linewidth]{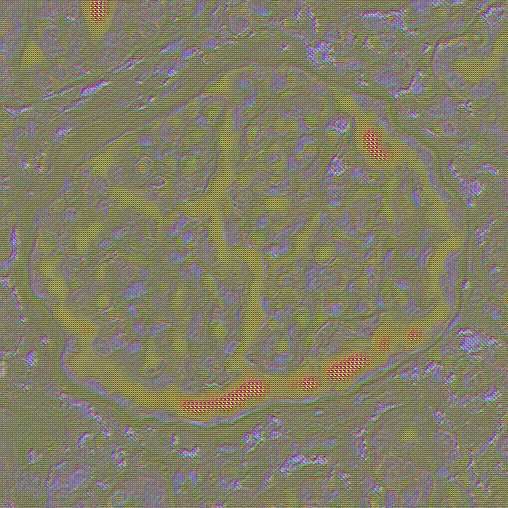} &
        \includegraphics[width=0.22\linewidth]{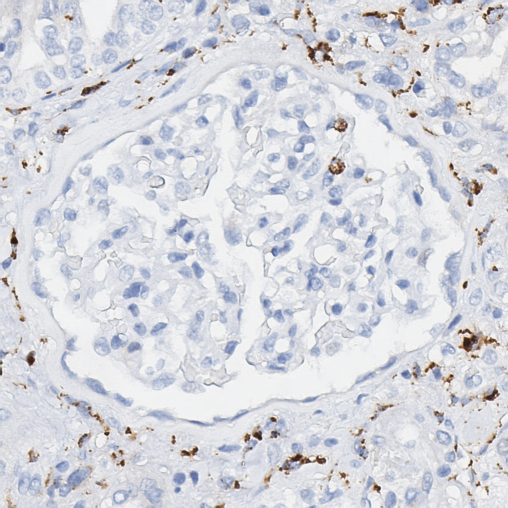}
    \end{tabular} 
    \caption{CycleGAN w/Extra-channels example. Translation relevant information is shifted to the extra-channels to minimise noise in the CD68$\rightarrow$PAS translation. However,  biologically relevant features that may be used by a segmentation model are also directed to the extra-channel.}
    \label{fig:extrachannels}
\end{figure}

CycleGAN w/Extra-channels
should  be most effective with the more challenging
HC$\leftrightarrow$IHC translation
and less so with straightforward HC translations, however, the opposite is observed. Moreover, it generalises poorly. It may be that certain biological features common to the source and target, and which should be present in the translation, are directed to the additional reconstruction channel, as shown in Fig.\ \ref{fig:extrachannels}.
CycleGAN w/Self-supervision and DSA-CycleGAN achieve better IHC performance as these stains are most likely to be affected by noise.
However, CycleGAN w/Self-supervision's loss is not applied to the T$\rightarrow$PAS translation, a limitation that DSA-CycleGAN overcomes. The ablation study highlights the need to apply a loss to all parts of the CycleGAN, as only the full model (DSA-CycleGAN) increases performance.  DSA-CycleGAN's constraining of T$\rightarrow$PAS also leads to better generalisation to unseen data.

In summary, CycleGAN w/Gaussian-noise and CycleGAN w/Extra-channel are recommended when translating biologically similar stains, and CycleGAN w/Self-supervision and DSA-CycleGAN otherwise.

\section{Conclusions} 
\label{conclusion}
This article presented DSA-CycleGAN, an approach to stain translation that reduces noise during translation, 
that increased glomeruli segmentation performance across various target stains compared to CycleGAN. It has also explored several other approaches to prevent such noise.
Since DSA-CycleGAN introduces the possibility of constraining the T$\rightarrow$S translation path, it outperformed these, particularly when translating between `information-rich' and `information-poor' stains.
This allows the  loss to be used in a wider range of translation approaches, e.g.\ StarGAN \citep{vasiljevic2023histostargan}, encoder/decoder models \citep{Yang21}, etc, as it is not limited to data originating from, and translated to, the source stain.

\section{Compliance with ethical standards}
\label{sec:ethics}
Study performed in line with the principles of the Declaration of Helsinki. Approval granted by the Ethics Committee of Hanover Medical School (Date 12/07/2015, No.\ 2968-2015).

\section{Acknowledgments}
\label{sec:acknowledgments}
Funded by ANR HistoGraph (ANR-23-CE45-0038) and ArtIC project ``Artificial Intelligence for Care'' (ANR-20-THIA-0006-01), co-funded by \textit{Région Grand Est}, Inria Nancy - Grand Est, IHU Strasbourg, University of Strasbourg \& University of Haute-Alsace. We acknowledge the ERACoSysMed \& e:Med initiatives by BMBF, SysMIFTA (managed by PTJ, FKZ 031L-0085A; ANR, grant ANR-15-CMED-0004), Prof.\ C\'{e}dric Wemmert, and the team at MHH for the high-quality images \& annotations: N.\ Kroenke, N.\ Schaadt, V.\ Volk \& J.\ Schmitz. We thank Nvidia, the \textit{Centre de Calcul} (University of Strasbourg) \& GENCI-IDRIS (grant 2020-A0091011872) for GPU access.

\bibliographystyle{IEEEbib}
\bibliography{bibliography}

\end{document}